\documentclass[runningheads]{llncs}
\usepackage{graphicx}
\usepackage{floatrow}
\usepackage{comment}
\usepackage{amsmath,amssymb} 
\usepackage{color}
\usepackage{epstopdf}
\newfloatcommand{figurebox}{figure}[\nocapbeside][\dimexpr(\textwidth-\columnsep)/2\relax]
\newfloatcommand{tablebox}{table}[\nocapbeside][\dimexpr(\textwidth-\columnsep)/2\relax]

\usepackage{diagbox}

\usepackage{array}
\newcolumntype{C}[1]{>{\centering\arraybackslash}m{#1}}
\newcolumntype{R}[1]{>{\raggedleft\arraybackslash}m{#1}}
\newcolumntype{P}[1]{>{\raggedright\arraybackslash}p{#1}}
\newcolumntype{M}[1]{>{\centering\arraybackslash}m{#1}}
\usepackage{multirow}

\floatstyle{ruled}

\usepackage[symbol]{footmisc}

\newcommand{\etal}{\textit{et al}.}
\newcommand{\ie}{\textit{i}.\textit{e}. }
\newcommand{\eg}{\textit{e}.\textit{g}. }

\begin{document}
\pagestyle{headings}
\mainmatter
\def\ECCVSubNumber{2191}  

\title{Self-supervising Fine-grained Region Similarities\\ for Large-scale Image Localization} 

\titlerunning{Self-supervising Fine-grained Region Similarities}
%
\author{Yixiao Ge\inst{1}
\and
Haibo Wang\inst{3}
\and 
Feng Zhu\inst{2} \and 
Rui Zhao\inst{2} \and 
Hongsheng Li\inst{1}}
\authorrunning{Y. Ge et al.}
%
\institute{
The Chinese University of Hong Kong \and 
SenseTime Research \and 
China University of Mining and Technology \\
\email{\{yxge@link,hsli@ee\}.cuhk.edu.hk} \\
\email{\{zhufeng,zhaorui\}@sensetime.com}  \quad
\email{haibo@cumt.edu.cn}
}
\maketitle

\begin{abstract}
The task of large-scale retrieval-based image localization is to estimate the geographical location of a query image by recognizing its nearest reference images from a city-scale dataset. However, the general public benchmarks only provide noisy GPS labels associated with the training images, which act as weak supervisions for learning image-to-image similarities. Such label noise prevents deep neural networks from learning discriminative features for accurate localization. To tackle this challenge, we propose to self-supervise image-to-region similarities in order to fully explore the potential of difficult positive images alongside their sub-regions. The estimated image-to-region similarities can serve as extra training supervision for improving the network in generations, which could in turn gradually refine the fine-grained similarities to achieve optimal performance. Our proposed self-enhanced image-to-region similarity labels effectively deal with the training bottleneck in the state-of-the-art pipelines without any additional parameters or manual annotations in both training and inference. Our method outperforms state-of-the-arts on the standard localization benchmarks by noticeable margins and shows excellent generalization capability on multiple image retrieval datasets.\footnote[2]{Code of this work is available at \url{https://github.com/yxgeee/SFRS}.}
\end{abstract}

\section{Introduction}

\begin{figure}[t]
    \centering
    \includegraphics[width=1.0\linewidth]{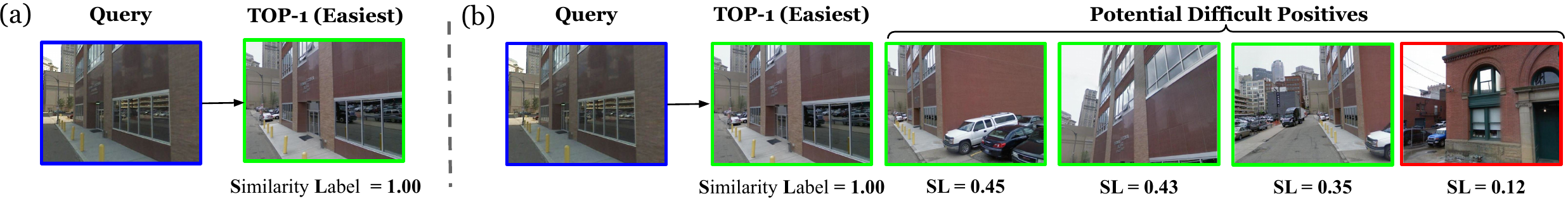}
    \caption{
    (a) To mitigate the noise with weak GPS labels, existing works \cite{arandjelovic2016netvlad,liu2019stochastic} only utilized the easiest top-$1$ image of the query for training.
    (b) We propose to adopt self-enhanced similarities as {soft} training supervisions to effectively explore the potential of difficult positives}
    \label{fig:fig1}
\end{figure}
\begin{figure}[t]
    \centering
    \includegraphics[width=1.0\linewidth]{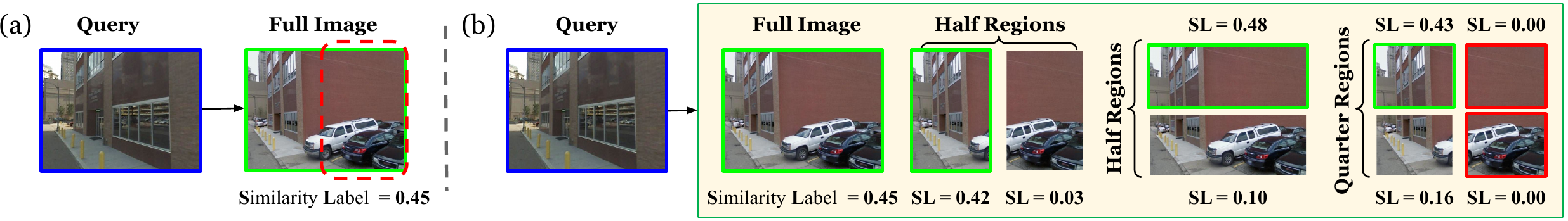}
    \caption{
    (a) Even with true positive pairs, the image-level supervision provides misleading information for non-overlapping regions.
    (b) We further estimate fine-grained soft labels to refine the supervision by measuring query-to-region similarities}
    \label{fig:fig2}
\end{figure}

Image-based localization (IBL) aims at estimating the location of a given image by identifying reference images captured at the same places from a geo-tagged database.
The task of IBL has long been studied since the era of hand-crafted features \cite{lowe2004distinctive,philbin2007object,arandjelovic2013all,jegou2010aggregating,jegou2011aggregating,perronnin2010large} and has been attracting increasing attention with the advances of convolutional neural networks (CNN) \cite{zeiler2014visualizing}, motivated by its wide applications in SLAM \cite{mur2015orb,hane20173d} and virtual/augmented reality \cite{castle2008video}.
Previous works have been trying to tackle IBL as image retrieval \cite{arandjelovic2016netvlad,liu2019stochastic,kim2017learned}, 2D-3D structure matching \cite{liu2017efficient,sattler2011fast} or geographical position classification \cite{vo2017revisiting,weyand2016planet,hongsuck2018cplanet} tasks.
In this paper, we treat the problem as an image retrieval task, given its effectiveness and feasibility in large-scale and long-term localization.

The fundamental challenge faced by image retrieval-based methods \cite{arandjelovic2016netvlad,liu2019stochastic,kim2017learned} is to learn image representations that are discriminative enough to tell apart repetitive and similar-looking locations in GPS-tagged datasets.
It is cast as a weakly-supervised task because geographically close-by images may not depict the same scene when facing different directions.

To avoid being misled by noisy and weak GPS labels, most works \cite{arandjelovic2016netvlad,liu2019stochastic} utilized on-the-fly first-ranking images of the queries as their positive training samples, \ie to force the queries to be closer to their already nearest neighbors (Fig. \ref{fig:fig1}(a)). Consequently, a paradox arises when only the most confident, or in other words, the \emph{easiest} on-the-fly positives are utilized in the training process, but these images in turn result in a lack of robustness to varying conditions as the first-ranking images for queries might be too simple to provide enough supervisions for learning robust feature representations. To tackle the issue, we argue that difficult positives are needed for providing informative supervisions.

Identifying the true difficult positives, however, is challenging with only geographical tags provided from the image database. Therefore, the network is easy to collapse when na\"ively trained with lower-ranking positive samples, which might not have overlapping regions with the queries at all. Such false positives might deteriorate the feature learning. Kim \etal \cite{kim2017learned} attempted to mine true positive images by verifying their geometric relations, but it is limited by the accuracy of off-the-shelf geometric techniques \cite{hartley2003multiple,fischler1981random}. In addition, even if pairs of images are indeed positive pairs, they might still have non-corresponding regions. The correct image-level labels might not necessarily be the correct region-level labels. Therefore, the ideal supervisions between paired positive images should provide region-level correspondences for more accurately supervising the learning of local features.

To tackle the above mentioned challenge,  we propose to estimate informative \emph{soft} image-to-region supervisions for training image features with noisy positive samples in a self-supervised manner. 
In the proposed system, an image retrieval network is trained in generations, which gradually improves itself with self-supervised image-to-region similarities. We train the network for the first generation following the existing pipelines \cite{arandjelovic2016netvlad,liu2019stochastic}, after which the network can be assumed to successfully capture most feature distribution of the training data. 
For each query image, however, its $k$-nearest gallery images according to 
the learned features might still contain noisy (false) positive samples. Directly utilizing them as difficult true positives in existing pipelines might worsen the learned features. We therefore propose to utilize their previous generation's query-gallery similarities to serve as the \emph{soft} supervision for training the network in a new generation. The generated soft supervisions are gradually refined and sharpened as the network generation progresses. In this way, the system is no longer limited to learning only from the on-the-fly \emph{easiest} samples, but can fully explore the potential of the difficult positives and mitigate their label noise with refined soft confidences (Fig. \ref{fig:fig1}(b)).

However, utilizing only the image-level soft supervisions for paired positive images simply forces features from all their spatial regions to approach the same target scores (Fig. \ref{fig:fig2}(a)). 
Such an operation would hurt the network's capability of learning discriminative local features.
To mitigate this problem, we further decompose the matching gallery images into multiple sub-regions of different sizes. The query-to-region similarities estimated from the previous generation's network can serve as the refined soft supervisions for providing more fine-grained guidance for feature learning (Fig. \ref{fig:fig2}(b)). The above process can iterate and train the network for multiple generations to progressively provide more accurate and fine-grained image-to-region similarities for improving the features.

Our contributions can be summarised as three-fold.
(1) We propose to estimate and self-enhance the image similarities of top-ranking gallery images to fully explore the potential of difficult positive samples in image-based localization (IBL). The self-supervised similarities serve as refined training supervisions to improve the network in generations, which in turn, generates more accurate image similarities.
(2) We further propose to estimate image-to-region similarities to
provide region-level supervisions for enhancing the learning of local features.
(3) The proposed system outperforms state-of-the-art methods on standard localization benchmarks by noticeable margins, and shows excellent generalization capability on both IBL and standard image retrieval datasets.

\section{Related Work}

\noindent\textbf{Image-based Localization (IBL).}
Existing works on image-based localization can be grouped into image retrieval-based \cite{arandjelovic2016netvlad,liu2019stochastic,kim2017learned}, 2D-3D registration-based \cite{liu2017efficient,sattler2011fast} and per-position classification-based \cite{vo2017revisiting,weyand2016planet,hongsuck2018cplanet} methods.
Our work aims at solving the training bottleneck of the weakly-supervised image retrieval-based IBL problem. We therefore mainly discuss related solutions that cast IBL as an image retrieval task.
Retrieval-based IBL is mostly related to the traditional place recognition task \cite{arandjelovic2014dislocation,chen2011city,knopp2010avoiding,schindler2007city,torii201524}, which has long been studied since the era of hand-engineered image descriptors, \eg SIFT \cite{lowe2004distinctive}, BoW \cite{philbin2007object}, VLAD \cite{arandjelovic2013all} and Fisher Vector \cite{perronnin2010large}.
Thanks to the development of deep learning \cite{lecun2015deep},
NetVLAD \cite{arandjelovic2016netvlad} successfully transformed dense CNN features for localization by proposing a learnable VLAD layer to effectively aggregate local descriptors with learnable semantic centers.
Adopting the backbone of NetVLAD,
later works further looked into multi-scale contextual information \cite{kim2017learned} or effective metric learning \cite{liu2019stochastic} to achieve better performance.
Our work has the similar motivation with \cite{kim2017learned},
\ie to mine difficult positive samples for more effective local feature learning.
However, 
\cite{kim2017learned} was only able to roughly mine positive images by adopting off-the-shelf geometric verification techniques \cite{hartley2003multiple,fischler1981random} and required noticeable more parameters for both training and inference.
In contrast, our method self-enhances fine-grained supervisions by gradually estimating and refining image-to-region soft labels without any additional parameters.


\noindent\textbf{Self-supervised label estimation} has been widely studied in self-supervised learning methods \cite{goyal2019scaling,kolesnikov2019revisiting,doersch2015unsupervised,noroozi2017representation,paulin2015local}, where the network mostly learns to predict a collective label set by properly utilizing the capability of the network itself.
Some works \cite{caron2018deep,ge2020mutual,ge2020selfpaced,ge2020structured} proposed to create task-relative pseudo labels,
\eg \cite{caron2018deep} generated image-level pseudo labels for classification by clustering features on-the-fly.
The others \cite{noroozi2016unsupervised,gidaris2018unsupervised,ge2018fd,zhou2020rotate} attempted to optimize the network by dealing with pretext tasks whose labels are more easily to create, \eg
\cite{noroozi2016unsupervised} solved jigsaw puzzles and 
\cite{gidaris2018unsupervised} predicted the rotation of transformed images.
Inspired by the self-supervised learning methods,
we propose a self-supervised image-to-region label creation scheme for training the network in generations, which effectively
mitigates the image-to-image label noise caused by weak geo-tagged labels.

\noindent\textbf{Self-distillation.}
Training in generations via self-predicted soft labels has been investigated in self-distillation methods \cite{furlanello2018born,zhang2020discriminability,xie2020self}. However, soft labels in these methods are not applicable to our task, as they focus on the classification problem with predefined classes. We successfully generalize self-distillation to the weakly-supervised IBL problem by proposing to generate soft labels for both query-to-gallery and query-to-region similarities in the retrieval task.

\section{Method}

\begin{figure}[th]
    \centering
    \includegraphics[width=0.92\linewidth]{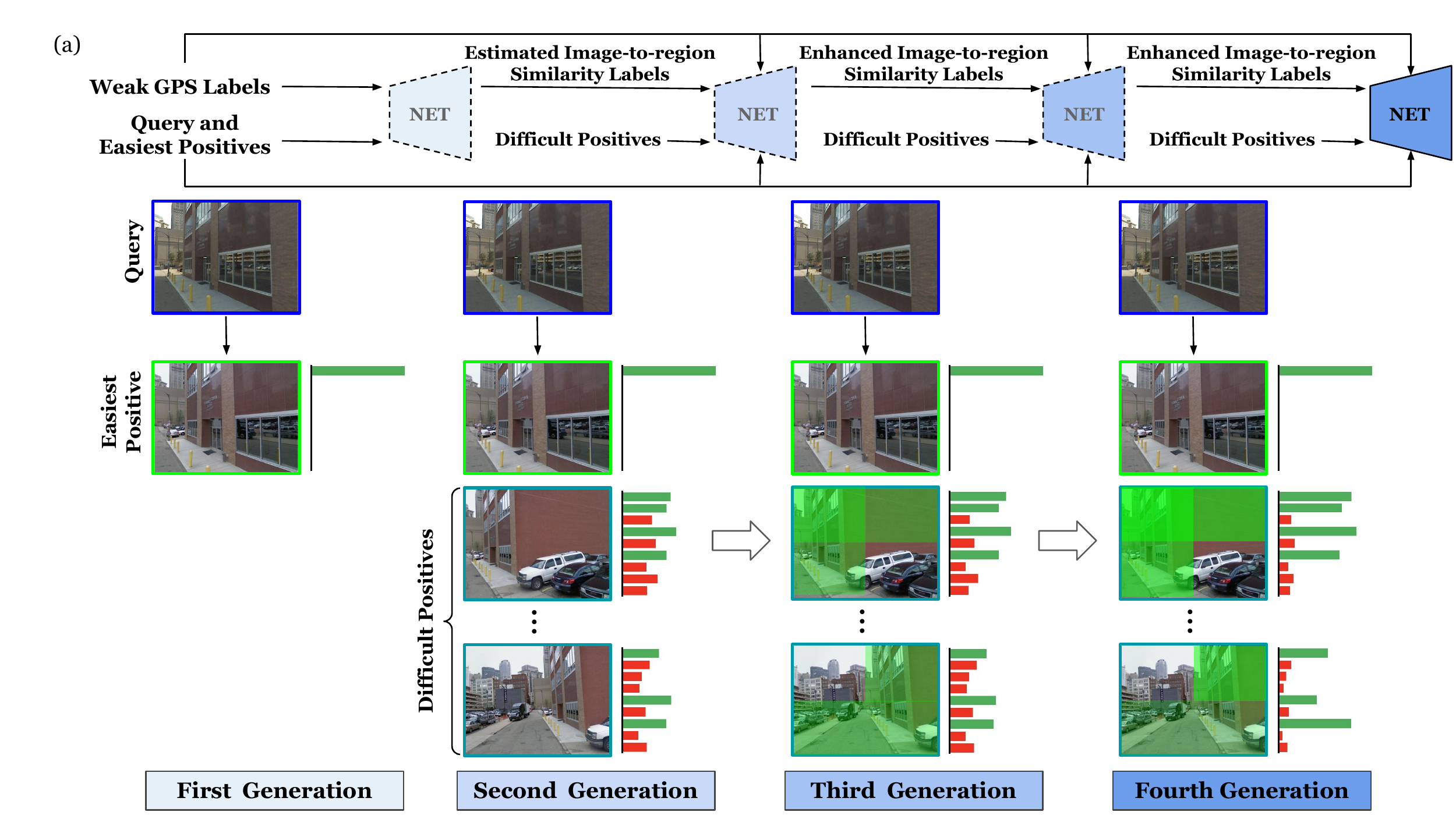} \\
    \includegraphics[width=0.9\linewidth]{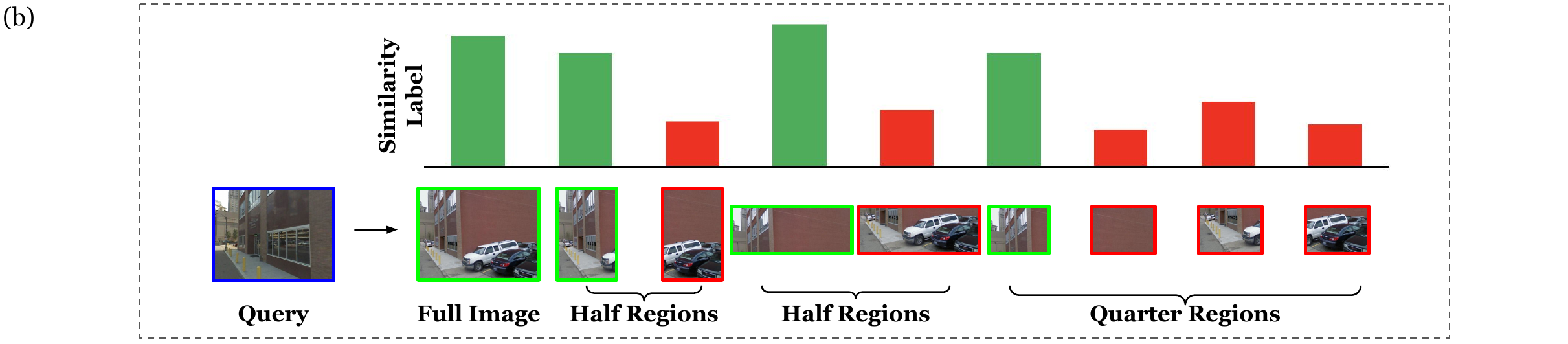} \\
     \caption{Illustration of our proposed self-supervised image-to-region similarities for image localization, where the image-to-region similarities are gradually refined via training the network in generations as shown in (a). 
    The sidebars for each difficult positive image demonstrate the soft similarity labels for the full image, left half, right half, top half, bottom half, top-left quarter, top-right quarter, bottom-left quarter and bottom-right quarter sub-regions respectively, detailed in (b), whereas the easiest positives only have one bar indicating the similarity label for the full image. Note that the most difficult negative samples utilized for joint training are not shown in the figure for saving space}
    \label{fig:fm}
\end{figure}

We propose to self-supervise image-to-region similarities for tackling the problem of noisy pairwise image labels in image-based localization (IBL). In major public benchmarks, each training image is generally associated with a GPS location tag. However, images geographically close to each other do not necessarily share overlapping regions. This is because the images might be captured from opposite viewing directions. The GPS tags could only help to geographically discover  potential positive images with much label noise.
Such a limitation causes bottleneck for effectively training the neural networks to recognize challenging positive pairs. There were previous methods \cite{arandjelovic2016netvlad,liu2019stochastic,kim2017learned} tackling this problem. 
However, 
they either ignored the potential of difficult but informative positive images, or required off-the-shelf time-consuming techniques with limited precision for filtering positives.

As illustrated in Fig. \ref{fig:fm}, the key of our proposed framework is to gradually enhance image-to-region similarities as the training proceeds, which can in turn act as soft and informative training supervisions for iteratively refining the network itself.
In this way, the potential of difficult positive samples can be fully explored and the network trained with such similarities can encode more discriminative features for accurate localization.

\subsection{Retrieval-based IBL Methods Revisit}
\label{sec:revisit}

State-of-the-art retrieval-based IBL methods \cite{arandjelovic2016netvlad,kim2017learned,liu2019stochastic} adopt the following general pipeline with different small modifications.
They generally train the network with learnable parameters $\theta$ by feeding triplets, each of which consists of one query image $q$, its {easiest} positive image $p^*$ and its multiple most difficult negative images $\{n_j |^N_{j=1} \}$.
Such triplets are sampled according to the image features $f_\theta$ encoded by the current network on-the-fly. 
The easiest positive image $p^*$ is the top-$1$ ranking gallery image within $10$ meters of the query $q$,
and the most difficult negative images $\{n_j |^N_{j=1} \}$ are randomly sampled from top-$1000$ gallery images that are more than $25$ meters away from $q$.

The network usually consists of a backbone encoder and a VLAD layer \cite{arandjelovic2016netvlad},
where the encoder embeds the image into a dense feature map and the VLAD layer further aggregates the feature map into a compact feature vector.
The network can be optimized with
a triplet ranking loss \cite{arandjelovic2016netvlad,kim2017learned} or a contrastive loss \cite{radenovic2016cnn}.
More recently, Liu \etal \cite{liu2019stochastic} proposed a softmax-based loss with dot products for better maximizing the ratio between the query-positive pair against multiple query-negative pairs,
which is formulated as
\begin{align} 
\mathcal{L}_{\mathrm{hard}}(\theta) = - \sum_{j=1}^{N} \log \frac{\exp \langle {f_\theta^q}, f_\theta^{p^*} \rangle  }
{\exp \langle {f_\theta^q}, f_\theta^{p^*} \rangle   + \exp \langle {f_\theta^q}, f^{n_j}_\theta \rangle },
\label{eq:baseline}
\end{align}
where 
$\theta$ is the network parameters, $f^q_\theta$ is the query image features encoded by the current network, and $\langle \cdot, \cdot \rangle$ denotes dot product between two vectors.
Trained by this pipeline, the network can capture most feature distributions and generate acceptable localization results.
Such a network acts as the \textbf{baseline} in our paper.

However,
an obvious problem arises: the network trained with the easiest positive images alone cannot well adapt to challenging positive pairs, as there generally exists variations in viewpoints, camera poses and focal lengths in real-world IBL datasets.
In addition,
the image-level weak supervisions provide misleading information for learning local features. There are non-overlapping regions even in true positive pairs. The network trained with image-level supervisions might pull the non-overlapping regions to be close to each other in the feature space. Such incorrect image-level supervisions impede the feature learning process.
To tackle such challenges,
we propose to utilize self-enhanced fine-grained supervisions for supervising both image-to-image and image-to-region similarities. In this way, difficult positive images as well as their sub-regions can be fully exploited to improve features for localization.

\subsection{Self-supervising Query-gallery Similarities}
\label{sec:image}

A na\"ive way to utilize difficult positive images is to directly train the network with lower-ranking positive images (\eg, top-$k$ gallery images) as positive samples with ``hard'' (one-hot) labels in Eq. \eqref{eq:baseline}.
However, this may make the training process deteriorate,
because the network cannot effectively tell the true positives apart from the false ones when trained with the existing pipeline in Sec. \ref{sec:revisit}.
To mitigate the label noise and make full use of the difficult positive samples,
we propose to self-supervise query-gallery similarities of top-ranking gallery images which also act as the refined soft supervisions, and to gradually improve the network with self-enhanced similarity labels in generations.

We use $\theta_\omega$ to indicate the network parameters in different generations, denoted as $\{ \theta_\omega |_{\omega=1}^\Omega \}$, where $\Omega$ is the total number of generations. 
In each generation, the network is {initialized} by 
the same ImageNet-pretrained \cite{deng2009imagenet} parameters
and is trained \textit{again} with new supervision set until convergence.
We adopt the same training process for each generation, except for the first initial generation,
for which we train the network parameter $\theta_1$
following the general pipeline in Sec. \ref{sec:revisit}.
Before proceeding to training the 2nd-generation network, the query-gallery feature distances are estimated by the 1st-generation network. For each query image $q$, its $k$-reciprocal nearest neighbors $p_1, \cdots, p_k$ according to Euclidean distances are first obtained from the gallery image set.
The query-gallery similarities can be measured by dot products and normalized by a softmax function with temperature $\tau_1$ over the top-$k$ images,
\begin{align}
\label{eq:S_i2i}
    \mathcal{S}_{\theta_1}(q,p_1,\cdots,p_k; \tau_1) &= \mathrm{softmax} \left( \left[ \langle {f_{\theta_1}^q}, f_{\theta_1}^{p_1} \rangle/\tau_1, \cdots, \langle {f_{\theta_1}^q}, f_{\theta_1}^{p_k} \rangle/\tau_1 \right]^\top \right),
\end{align}
where $f^{p_i}_{\theta_1}$ is the encoded feature representations of the $i$th gallery image by the 1st-generation network, and $\tau_1$ is temperature hyper-parameter that makes the similarity vector sharper or smoother. 

Rather than na\"ively treating all top-$k$ ranking images as the true positive samples for training the next-generation network, we propose to utilize the relative similarity vector $\mathcal{S}_{\theta_1}$ as extra training supervisions. There are two advantages of adopting such supervisions: (1) We are able to use the top-ranking gallery images as candidate positive samples. Those plausible positive images are more difficult and thus more informative than the on-the-fly easiest positive samples for learning scene features. (2) To mitigate the inevitable label noise from the plausible positive samples, we propose to train the next-generation network to approach the relative similarity vectors $\mathcal{S}_{\theta_1}$, which ``softly'' measures the relative similarities between $(q, p_1), \cdots, (q, p_k)$ pairs. At earlier generations, we set the temperature parameter $\tau_\omega$ to be large, as the relative similarities are less accurate. The large temperature makes the similarity vector $\mathcal{S}_{\theta_\omega}$ more equally distributed. At later generations, the network becomes more accurate. The relative similarity vector $\mathcal{S}_{\theta_\omega}$'s maximal response is more accurate to identify the true positive samples. Lower temperatures are used to make $\mathcal{S}_{\theta_\omega}$ concentrate on a small number of gallery images.

The relative similarity vectors $\mathcal{S}_{\theta_1}$ estimated by the 1st-generation network are used to supervise the 2nd-generation network via a ``soft'' cross-entropy loss,
\begin{align}
\label{eq:soft}
    \mathcal{L}_\mathrm{soft}(\theta_2) = \ell_{ce} \left( \mathcal{S}_{\theta_2}(q,p_1,\cdots,p_k;1), \mathcal{S}_{\theta_1}(q,p_1,\cdots,p_k;\tau_1)\right),
\end{align}
where $\ell_{ce}(y,\hat{y})=- \sum_i \hat{y}(i)  \log ({y}(i))$ denotes the cross-entropy loss. Note that only the learning target $\mathcal{S}_{\theta_1}$ adopts the temperature hyper-parameter to control its sharpness. 
In our experiments, the performance of the trained network generally increases with generations and saturates around the 4th generation, with temperatures set as 0.07, 0.06, 0.05, respectively.
In each iteration, both the ``soft'' cross-entropy loss and the original hard loss (Eq. \eqref{eq:baseline}) are jointly adopted as ${\cal L}(\theta) = {\cal L}_\textrm{hard}(\theta) + \lambda {\cal L}_\mathrm{soft}(\theta)$ for supervising the feature learning.

By self-supervising query-gallery similarities, we could fully utilize the difficult positives obtained from the network of previous generations. The ``soft'' supervisions can effectively mitigate the label noise.
However, even with the refined labels, paired positive images' local features from all their spatial regions are trained to approach the same similarity scores, which provide misleading supervision for those non-overlapping regions. The network's capability on learning discriminative local features is limited given only such image-level supervisions.

\subsection{Self-supervising Fine-grained Image-to-region Similarities}
\label{sec:i2r}

To tackle the above challenge,
we propose to fine-grain the image-to-image similarities into image-to-region similarities to generate more detailed supervisions.
Given the top-$k$ ranking images $p_1, \cdots, p_k$ for each query $q$, instead of directly calculating image-level relative similarity vector $\mathcal{S}_{\theta_\omega}$, we can further decompose each plausible positive image's feature maps into 4 half regions (top, bottom, left, right halves) and 4 quarter regions (top-left, top-right, bottom-left, bottom-right quarters).
Specifically,
The gallery image $p_i$'s feature maps $m_{i}$ are first obtained by the network backbone before the VLAD layer.
We split $m_i$ into 4 half regions and 4 quarter regions, and then feed them into the aggregation VLAD respectively for obtaining 8 region feature vectors $\{f_\theta^{r_i^1}, f_\theta^{r_i^2}, \cdots, f_\theta^{r_i^8}\}$ corresponding to the 8 image sub-regions $\{{r_i^1}, {r_i^2}, \cdots, {r_i^8}\}$, each of which depicts the appearance of one sub-region of the gallery scene.
Thus, the query-gallery supervisions are further fine-grained by measuring the relative query-to-region similarities,
\begin{align}
\label{eq:S_i2r}
    \mathcal{S}^r_{\theta_\omega}(q,p_1,\cdots,p_k; &\tau_\omega) = \mathrm{softmax} \left( \big[ \langle {f_{\theta_\omega}^q}, f_{\theta_\omega}^{p_1} \rangle/\tau_\omega, \langle {f_{\theta_\omega}^q}, f_{\theta_\omega}^{r^1_1} \rangle/\tau_\omega, \cdots, \langle {f_{\theta_\omega}^q}, f_{\theta_\omega}^{r^8_1} \rangle/\tau_\omega, \right. \nonumber \\ 
     &\cdots, \left. \langle {f_{\theta_\omega}^q}, f_{\theta_\omega}^{p_k} \rangle/\tau_\omega, \langle {f_{\theta_\omega}^q}, f_{\theta_\omega}^{r^1_k} \rangle/\tau_\omega, \cdots, \langle {f_{\theta_\omega}^q}, f_{\theta_\omega}^{r^8_k} \rangle/\tau_\omega \big] \right) .
\end{align}
The ``soft'' cross-entropy loss in Eq. \eqref{eq:soft} can be extended such that it can learn from the fine-grained similarities in Eq. \eqref{eq:S_i2r} to encode more discriminative local features,
\begin{align}
\label{eq:soft_r}
    \mathcal{L}_\mathrm{soft}^r(\theta_\omega) = \ell_{ce} \left( \mathcal{S}^r_{\theta_\omega}(q,p_1,\cdots,p_k;1), \mathcal{S}^r_{\theta_{\omega-1}}(q,p_1,\cdots,p_k;\tau_{\omega-1}) \right).
\end{align}

The image-to-region similarities can also be used for mining the most difficult negative gallery-image regions for each query $q$. The mined negative regions can be used in the hard loss in Eq. \eqref{eq:baseline}.
For each query $q$, the accurate negative images could be easily identified by the geographical GPS labels.
We propose to further mine the most difficult region $n^*_j$ for each of the negative images $n_j$ on-the-fly by measuring query-to-region similarities with the current network. 
The image-level softmax-based loss in Eq. \eqref{eq:baseline} could be refined with the most difficult negative regions as
\begin{align} 
\mathcal{L}_{\mathrm{hard}}^r(\theta_\omega) = - \sum_{j=1}^{N} \log \frac{\exp \langle {f_{\theta_\omega}^q}, f_{\theta_\omega}^{p^*} \rangle  }
{\exp \langle {f_{\theta_\omega}^q}, f_{\theta_\omega}^{p^*} \rangle   +  \exp \langle {f_{\theta_\omega}^q}, f^{n^*_j}_{\theta_\omega} \rangle }.
\label{eq:hard_r}
\end{align}
Note that the image-to-region similarities for selecting the most difficult negative regions are measured by the network on-the-fly, while those for supervising difficult positives in Eq. \eqref{eq:soft_r} are measured by the previous generation's network.

In our proposed framework, the network is trained with extended triplets, each of which consists of one query image $q$, the easiest positive image $p^*$, the difficult positive images $\{p_i|^k_{i=1}\}$, and the most difficult regions in the negative images $\{n_j^*|^N_{j=1}\}$.
The overall objective function of generation $\omega$ is formulated as
\begin{align}
    \mathcal{L}(\theta_\omega) = \mathcal{L}^r_\mathrm{hard}(\theta_\omega) + \lambda  \mathcal{L}^r_\mathrm{soft}(\theta_\omega),
    \label{eq:all}
\end{align}
where $\lambda$ is the loss weighting factor. The multi-generation training can be performed similarly to that mentioned in Sec. \ref{sec:image} to self-supervise and to gradually refine the image-to-region similarities.

\subsection{Discussions}

\noindent\textbf{Why not decompose queries for self-supervising region-to-region similarities?}
We observe that region-to-region similarities might contain many superficially easy cases that are not informative enough to provide discriminative training supervisions for the network. For instance, the similarity between the sky regions of two overlapping images might be too easy to serve as effective training samples.
Image-to-region similarities could well balance the robustness and granularity of the generated supervisions, as region-to-region similarities have more risk to be superficial.

\noindent\textbf{Why still require $\mathcal{L}_\text{hard}^r$?}
We observe that the top-$1$ gallery images obtained from the $k$-reciprocal nearest neighbors can almost always act as true positives for training with hard supervisions in Eq. \eqref{eq:hard_r}.
The hard loss could stabilize the training process to avoid error amplification.


\section{Experiments}

\subsection{Implementation Details}

\noindent\textbf{Datasets.}
We utilize the Pittsburgh benchmark dataset \cite{torii2013visual} for optimizing our image retrieval-based localization network following the experimental settings of state-of-the-art methods \cite{liu2019stochastic,arandjelovic2016netvlad}.
Pittsburgh consists of a large scale of panoramic images captured at different times and are associated with noisy GPS locations.
Each panoramic image is projected to create 24 perspective images.
For fair comparison,
we use the subset Pitts30k for training and select the best model that achieves the optimal performance on the val-set of Pitts30k.
The Pitts30k-train contains 7,416 queries and 10,000 gallery images, 
and the Pitts30k-val consists of 7,608 queries and 10,000 gallery images.
We obtain the final retrieval results by ranking images in the large-scale Pitts250k-test,
which contains 8,280 probes and 83,952 database images.

To verify the generalization ability of our method on different IBL tasks,
we directly evaluate our models trained on Pitts30k-train on the Tokyo 24/7 \cite{torii201524} dataset,
which is quite challenging since the queries were taken in varying conditions.
In addition, we evaluate the model's generalization ability on standard image retrieval datasets, \eg the Oxford 5k \cite{philbin2007object}, Paris 6k \cite{Philbin08}, and Holidays \cite{jegou2008hamming}.

\noindent\textbf{Evaluation.}
During inference, we perform PCA whitening whose parameters are learnt on the Pitts30k-train, reducing the feature dimension to 4,096.
We follow the same evaluation metric of \cite{liu2019stochastic,arandjelovic2016netvlad},
where the top-$k$ recall is measured on the localization datasets, Pitts250k-test \cite{torii2013visual} and Tokyo 24/7 \cite{torii201524}. The query image is determined to be successfully retrieved from top-$k$ if at least one of the top $k$ retrieved reference images locates within $d=25$ meters from the query image.
As for the image-retrieval datasets, Oxford 5k \cite{philbin2007object}, Paris 6k \cite{Philbin08}, and Holidays \cite{jegou2008hamming}, the mean average precision (mAP) is adopted for evaluation.

\noindent\textbf{Architecture.} 
For fair comparison, we adopt the same architecture used in \cite{liu2019stochastic,arandjelovic2016netvlad}, which comprises of a VGG-16 \cite{simonyan2014very} backbone and a VLAD layer \cite{arandjelovic2016netvlad} for encoding and aggregating feature representations. We use the ImageNet-pretrained \cite{deng2009imagenet} VGG-16 up to the last convolutional layer (\ie conv5) before ReLU, as the backbone. Following \cite{liu2019stochastic}, the whole backbone except the last convolutional block (\ie conv5) is frozen when trained on image-based localization datasets.

\noindent\textbf{Training details.} 
For data organization, each mini-batch contains $4$ triplets, each of which consists of one query image, one easiest positive image, top-$10$ difficult positive images and $10$ negative images.
The negative images are sampled following the same strategy in \cite{liu2019stochastic,arandjelovic2016netvlad}. Our model is trained by 4 generations, with 5 epochs in each generation. We empirically set the hyper-parameters $\lambda=0.5, \tau_1=0.07, \tau_2=0.06, \tau_3=0.05$ in all experiments. The stochastic gradient descent (SGD) algorithm is utilized to optimize the loss function, with momentum 0.9, weight decay 0.001, and a constant learning rate $=0.001$.

\subsection{Comparison with State-of-the-arts}

\begin{table}[t]
	\centering
	\caption{Comparison with state-of-the-arts on image-based localization benchmarks. Note that the network is only trained on Pitts30k-train and directly evaluated on both Tokyo 24/7 and Pitts250k-test datasets}
	\label{tab:sota}
	\begin{center}
	\begin{tabular}{P{3.7cm}|C{1.3cm}C{1.3cm}C{1.3cm}|C{1.3cm}C{1.3cm}C{1.3cm}}
	\hline
	\multicolumn{1}{c|}{\multirow{2}{*}{Method}} & \multicolumn{3}{c|}{Tokyo 24/7 \cite{torii201524}} & \multicolumn{3}{c}{Pitts250k-test \cite{torii2013visual}}  \\
	\cline{2-7}
	 \\[-2.3ex]
	\multicolumn{1}{c|}{} & R$@1$ & R$@5$ & R$@10$ & R$@1$ & R$@5$ & R$@10$ \\ 
	\hline \hline
    NetVLAD \cite{arandjelovic2016netvlad} (CVPR'16)& 73.3 & 82.9 & 86.0 & 86.0 & 93.2 & 95.1  \\
    CRN \cite{kim2017learned} (CVPR'17) & 75.2 & 83.8 & 87.3 & 85.5 & 93.5 & 95.5 \\
    SARE \cite{liu2019stochastic} (ICCV'19) & 79.7 & 86.7 & 90.5 & 89.0 & 95.5 & 96.8 \\
	\hline
	Ours & \textbf{85.4} & \textbf{91.1} & \textbf{93.3} & \textbf{90.7} & \textbf{96.4} & \textbf{97.6}  \\
	\hline
	\end{tabular}
	\end{center}
\end{table}

We compare with state-of-the-art image localization methods NetVLAD \cite{arandjelovic2016netvlad}, CRN \cite{kim2017learned} and SARE \cite{liu2019stochastic} on localization datasets Pitts250k-test \cite{torii2013visual} and Tokyo 24/7 \cite{torii201524} in this experiment. Our model is only trained on Pitts30k-train without any Tokyo 24/7 images. CRN (Contextual Reweighting Network) improves NetVLAD by proposing a contextual feature reweighting module for selecting most discriminative local features for aggregation. While a Stochastic Attraction and Repulsion Embedding (SARE) loss function is proposed on top of VLAD-aggregated feature embeddings in \cite{liu2019stochastic}. None of the above methods have well handled the bottleneck of weakly-supervised training for image localization.

Experimental results are shown in Tab. \ref{tab:sota}. The proposed method achieves 90.7\% rank-1 recall on Pitts250k-test, outperforming the second best 89.0\% obtained by SARE, with an improvement of 1.7\%. The feature embeddings learned by our method show very strong generalization capability on the challenging Tokyo 24/7 dataset where the rank-1 recall is significantly boosted to 85.4\%, up to 5.7\% performance improvement against SARE. The superior performances validate the effectiveness of our self-enhanced image-to-region similarities in learning discriminative features for image-based localization.

\begin{figure}[t]
    \centering
    \includegraphics[width=1\linewidth]{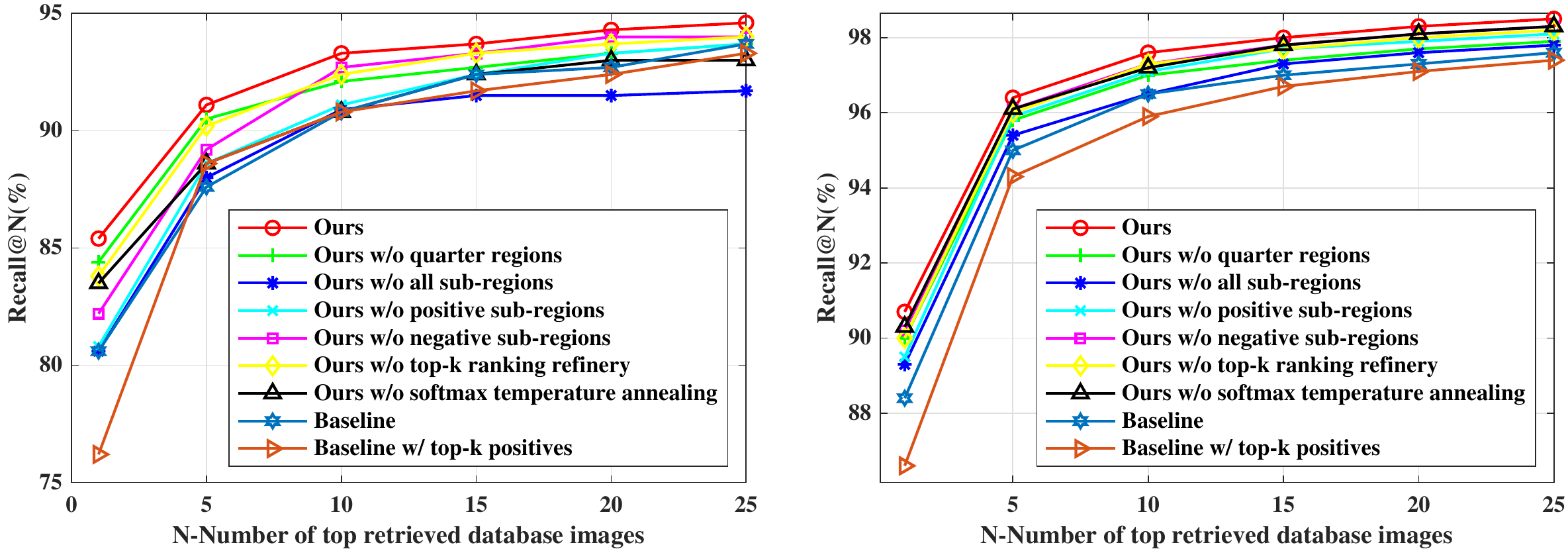} \\
     (a) Tokyo 24/7 \cite{torii201524} ~~~~~~~~~~~~~~~~~~~~~~~~~~~~ (b) Pitts250k-test \cite{torii2013visual}
    \caption{Quantitative results of ablation studies on our proposed method in terms of top-$k$ recall on Tokyo 24/7 and Pitts250k-test datasets. The models are only trained on Pitts30k-train set}
    \label{fig:fig_ablation}
\end{figure}

\subsection{Ablation Studies}

We perform ablation studies on Tokyo 24/7 \cite{torii201524} and Pitts250k-test \cite{torii2013visual} to analyse the effectiveness of the proposed method and shed light on the importance of supervisions from self-enhanced fine-grained image-to-region similarities. We illustrate quantitative results in Fig. \ref{fig:fig_ablation}, and show the detailed ablation experiments in Tab. \ref{tab:ablation}. ``Baseline'' is our re-implementation of SARE \cite{liu2019stochastic}, which is trained with only the best-matching positive image using $\mathcal{L}_{\mathrm{hard}}$ in Eq. (\ref{eq:baseline}). 

\noindent\textbf{Directly training with noisy difficult positives in the existing pipeline.} Existing image-based localization benchmarks only provide geo-tagged images for learning image-to-image similarities. These GPS labels are weak and noisy in finding true positive images for given query images. SARE \cite{liu2019stochastic} and previous works only choose the easiest positive for training, \ie the most similar one to the query in the feature space. Our proposed approach can effectively learn informative features from multiple difficult positives (top-$k$ positive images) by introducing self-enhanced image-region similarities as extra supervisions, thus boost the rank-1 recall to 90.7\% and 85.4\% on Pitts250k-test and Tokyo 24/7 respectively. 
To check whether such difficult (top-$k$) positive images can be easily used without our proposed self-enhanced image-to-region similarities to improve the final performance, we conduct an ablation experiment by training the ``Baseline'' with extra difficult positive images. This model is denoted as ``Baseline w/ top-$k$ positives'' in Tab. \ref{tab:ablation}. It shows that na\"ively adding more difficult positive images for training causes drastic performance drop, where the rank-1 recall decreases from 80.6\% (``Baseline'') to 76.2\% on Tokyo 24/7. The trend is similar on Pitts250k-test. This phenomenon validates the effectiveness of our self-enhanced image-to-region similarities  from difficult positive samples on the IBL tasks.

\noindent\textbf{Effectiveness of fine-grained image-to-region similarities.} As described in Sec. \ref{sec:i2r}, our proposed framework explores image-to-region similarities by dividing a full image into 4 half regions and 4 quarter regions. Such design is critical in our framework as query and positive images are usually only partially overlapped due to variations in camera poses. Simply forcing query and positive images to be as close as possible in feature embedding space regardless of non-overlapping regions would mislead feature learning, resulting in inferior performance. As shown in Tab. \ref{tab:ablation}, on Tokyo 24/7, the rank-1 recall drops from 85.4\% to 84.4\% (``Ours w/o quarter regions'') when the 4 quarter regions are excluded from training, and further drastically drops to 80.6\% (``Ours w/o all sub-regions'') when no sub-regions are used. The effectiveness of sub-regions in positive and negative images are compared by ``Ours w/o positive sub-regions'' and ``Ours w/o negative sub-regions'' in Tab. \ref{tab:ablation}. It shows that in both cases the rank-1 recall is harmed, and the performance drop is even more significant when positive sub-regions are ignored. The above observations indicate that difficult positives are critical for feature learning in IBL tasks, which have not been well investigated in previous works, while our fine-grained image-to-region similarities effectively help learn discriminative features from these difficult positives.

\begin{table}[t]
	\centering
	\caption{Ablation studies for our proposed method on individual components. The models are only trained on Pitts30k-train set}
	\label{tab:ablation}
	\begin{center}
	\begin{tabular}{P{6.5cm}|C{1.5cm}C{1.5cm}C{1.5cm}}
	\hline
	\multicolumn{1}{c|}{\multirow{2}{*}{Method}}  & \multicolumn{3}{c}{Tokyo 24/7 \cite{torii201524}} \\
	\cline{2-4}
	 \\[-2.3ex]
	\multicolumn{1}{c|}{} & R$@1$ & R$@5$ & R$@10$  \\ 
		\hline \hline
    Baseline  & 80.6 & 87.6 & 90.8 \\
    Baseline w/ top-$k$ positives & 76.2 & 88.6 & 90.8 \\
    Baseline w/ regions & 79.8 & 86.9 & 90.4  \\
    \hline
    Ours w/o all sub-regions & 80.6 & 88.0 & 90.9 \\
    Ours w/o positive sub-regions & 80.8 & 88.6 & 91.1 \\
    Ours w/o negative sub-regions & 82.2 & 89.2 & 92.7\\
    Ours w/o quarter regions & 84.4 & 90.5 & 92.1 \\
    \hline
    Ours w/o top-$k$ ranking refinery & 83.8 & 90.2 & 92.4 \\
    Ours w/o softmax temperature annealing & 83.5 & 88.6 & 90.8 \\
    \hline
	Ours & \textbf{85.4} & \textbf{91.1} & \textbf{93.3} \\
	\hline
	\end{tabular}
	\end{center}
\end{table}


\noindent\textbf{Effectiveness of self-enhanced similarities as soft supervisions.} When comparing ``Ours w/o all sub-regions'' and ``Baseline w/ top-$k$ positives'', one may find that the only difference between these two methods is $\mathcal{L}_{\mathrm{soft}}$ in Eq. (\ref{eq:soft}). By adding the extra objective $\mathcal{L}_{\mathrm{soft}}$ to ``Baseline w/ top-$k$ positives'', the rank-1 recall is significantly improved from 76.1\% to 80.6\% (the same with ``Baseline'') on Tokyo 24/7. ``Ours w/o all sub-regions'' even outperforms ``Baseline'' on Pitts250k-test as shown in Fig. \ref{fig:fig_ablation}. The above comparisons demonstrate the effectiveness of using our self-enhanced similarities as soft supervisions for learning from difficult positive images even at the full image level.

More importantly, soft supervisions serve as a premise for image-to-region similarities to work.
We evaluate the effects of soft supervisions at the region level, dubbed as ``Baseline w/ regions'', where the dataset is augmented with decomposed regions and only the noisy GPS labels are used. The result in Tab. \ref{tab:ablation} is even worse than ``Baseline'' since sub-regions with only GPS labels provide many too easy positives that are not informative enough for feature learning. 


\noindent\textbf{Benefit from top-$k$ ranking refinery.}
We propose to find $k$-reciprocal nearest neighbors of positive images for recovering more accurate difficult positives images for training in Sec. \ref{sec:image}. Although our variant (``Ours w/o top-$k$ ranking refinery'') without using $k$-reciprocal nearest neighbors refining top-$k$ images already outperforms ``Baseline'' by a large margin, ranking with $k$-reciprocal nearest neighbor further boosts the rank-1 recall by 1.6\% on Tokyo 24/7. The superior performance demonstrates that $k$-reciprocal nearest neighbor ranking can more accurately identify true positive images.

\noindent\textbf{Benefits from softmax temperature annealing.}
We validate the effectiveness of the proposed temperature annealing strategy (Sec. \ref{sec:image}) in this experiment by setting a constant temperature $\tau=0.07$ in all generations. Comparisons between ``Ours'' and ``Ours w/o softmax temperature annealing'' in Fig. \ref{fig:fig_ablation} and Tab. \ref{tab:ablation} show that temperature annealing is beneficial for learning informative features with our self-enhanced soft supervisions.

\begin{figure}[t]
    \centering
    \scriptsize
    \begin{center}
    \begin{tabular}{C{2.2cm}C{2.2cm}C{2.2cm}C{2.2cm}C{2.2cm}}
    \includegraphics[width=1.0\linewidth]{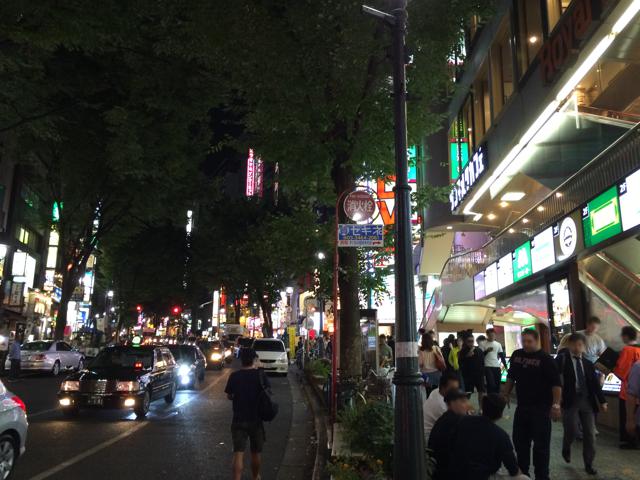} &
    \includegraphics[width=1.0\linewidth]{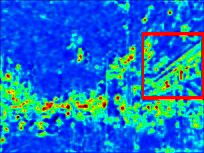} & \includegraphics[width=1.0\linewidth]{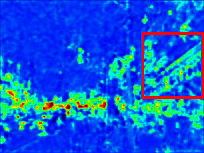} &
    \includegraphics[width=1.0\linewidth]{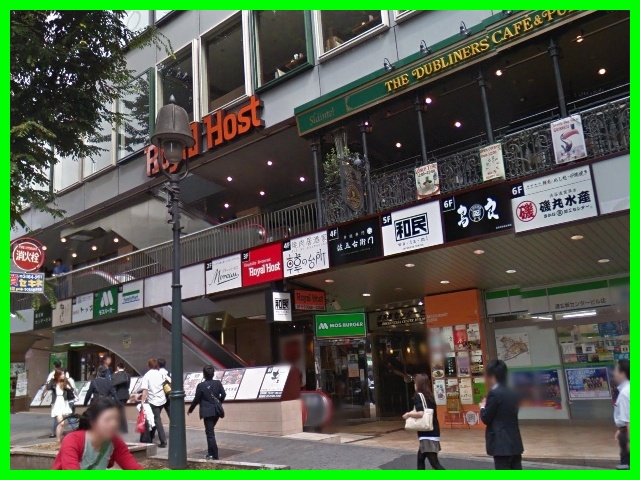} &
    \includegraphics[width=1.0\linewidth]{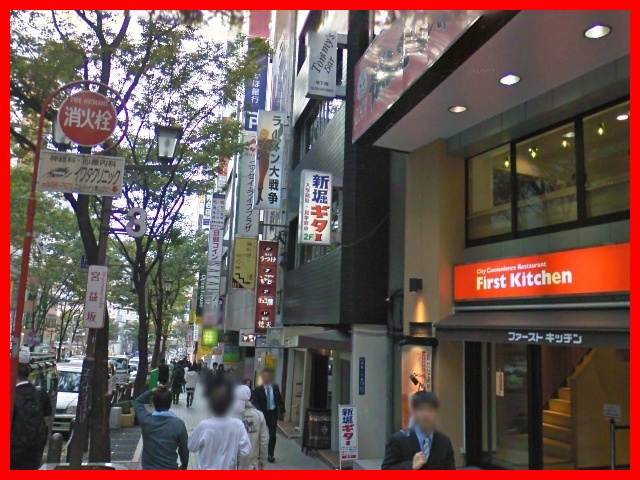} \\
    \includegraphics[width=1.0\linewidth]{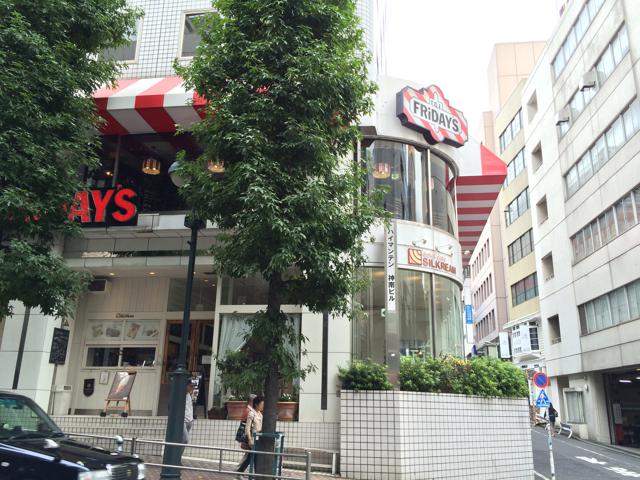} &
    \includegraphics[width=1.0\linewidth]{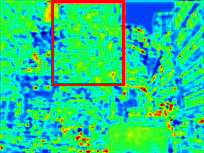} & \includegraphics[width=1.0\linewidth]{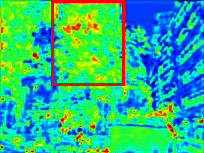} &
    \includegraphics[width=1.0\linewidth]{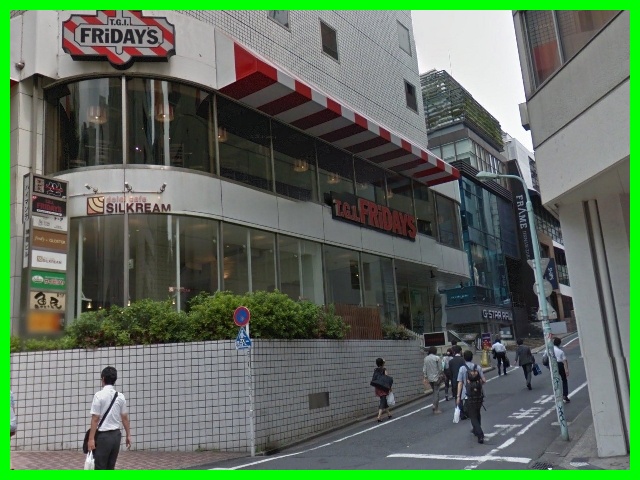} &
    \includegraphics[width=1.0\linewidth]{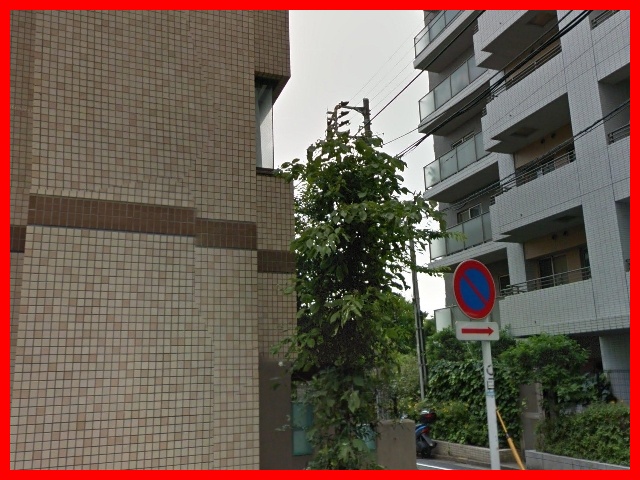} \\
    \includegraphics[width=1.0\linewidth]{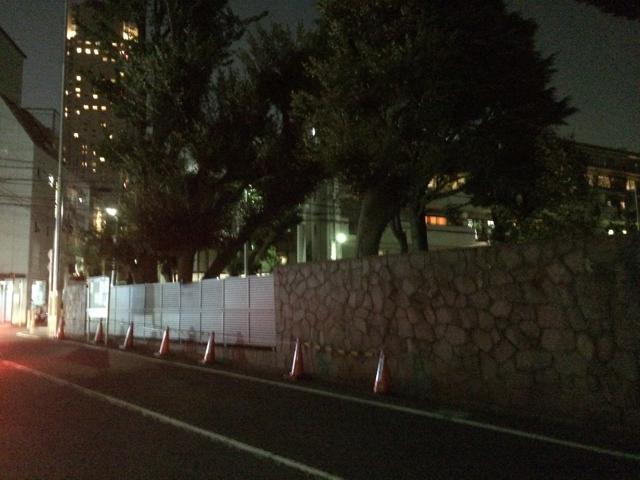} &
    \includegraphics[width=1.0\linewidth]{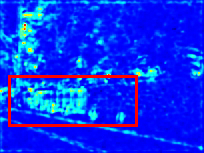} & \includegraphics[width=1.0\linewidth]{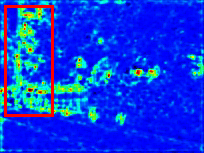} &
    \includegraphics[width=1.0\linewidth]{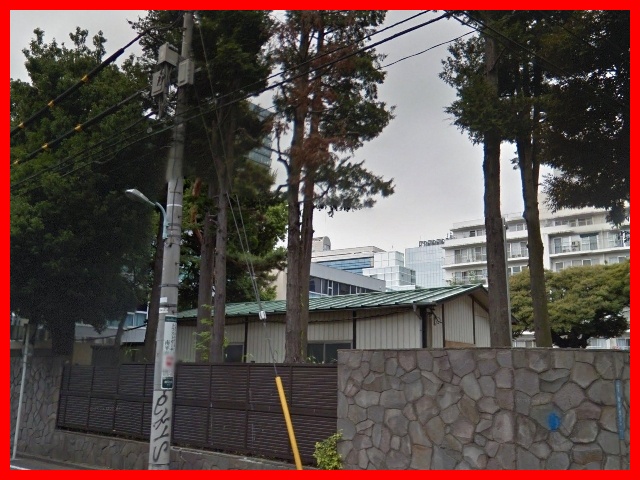} &
    \includegraphics[width=1.0\linewidth]{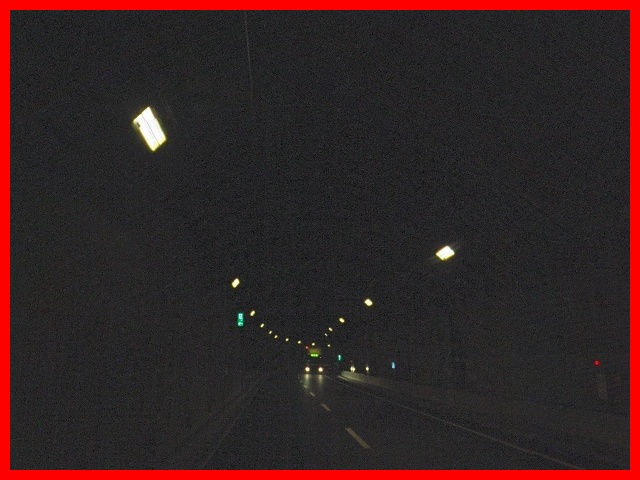} \\
    \centering (a) Query & (b) Our heatmap & (c) \cite{liu2019stochastic}'s heatmap & (d) Our top-$1$ & (e) \cite{liu2019stochastic}'s top-$1$ \\
    \end{tabular}
    \end{center}
    \caption{Retrieved examples of our method and state-of-the-art SARE \cite{liu2019stochastic} on Tokyo 24/7 dataset \cite{torii201524}. 
    The regions highlighted by red boxes illustrates the main differences.
     }
    \label{fig:imgs}
\end{figure}

\subsection{Qualitative Evaluation}

To better understand the superior performance of our method on the IBL tasks, we visualize the learned feature maps before VLAD aggregation as heatmaps, shown in Fig. \ref{fig:imgs}. 
In the first example, our method pays more attention on the discriminative shop signs than SARE, which provide valuable information for localization in city-scale street scenarios. In the second example, SARE incorrectly focuses on the trees, while our method learns to ignore such misleading regions by iteratively refining supervisions from fine-grained image-to-region similarities. Although both methods fail in the third example, our retrieved top-1 image is more reasonable with wall patterns similar to the query image. 

\begin{table}[t]
	\centering
	\caption{Evaluation on standard image retrieval datasets in terms of mAP ($\%$) to validate the generalization ability of the networks}
	\label{tab:natural}
	\begin{center}
	\begin{tabular}{P{3.7cm}|C{1.3cm}C{1.3cm}|C{1.3cm}C{1.3cm}|C{1.7cm}}
	\hline
	\multicolumn{1}{c|}{\multirow{2}{*}{Method}} & \multicolumn{2}{c|}{Oxford 5k \cite{philbin2007object}} & \multicolumn{2}{c|}{Paris 6k \cite{Philbin08}} & \multicolumn{1}{c}{\multirow{2}{*}{Holidays \cite{jegou2008hamming}}} \\
	\cline{2-5}
	 \\[-2.3ex]
	\multicolumn{1}{c|}{} & full & crop & full & crop & \multicolumn{1}{c}{} \\ 
	\hline \hline
    NetVLAD \cite{arandjelovic2016netvlad} (CVPR'16) & 69.1 & 71.6 & 78.5 & 79.7 & \textbf{83.1} \\
    CRN \cite{kim2017learned} (CVPR'17) & 69.2 & - & - & - & - \\
    SARE \cite{liu2019stochastic} (ICCV'19) & 71.7 & 75.5 & 82.0 & 81.1 & 80.7  \\
	\hline
	Ours & \textbf{73.9} & \textbf{76.7} & \textbf{82.5} & \textbf{82.4} & 80.5 \\
	\hline
	\end{tabular}
	\end{center}
\end{table}

\subsection{Generalization on Image Retrieval Datasets}

In this experiment, we evaluate the generalization capability of learned feature embeddings on standard image retrieval datasets by directly testing trained models without fine-tuning. 
The experimental results are listed in Tab. \ref{tab:natural}. 
``Full'' means the feature embeddings are extracted from the whole image, while only cropped landmark regions are used in the ``crop'' setting. 
Our method shows good generalization ability on standard image retrieval tasks, and outperforms other competitors on most datasets and test settings.
All compared methods do not perform well on Holidays due to the the fact that 
Holidays contains lots of natural sceneries, which are difficult to find in our street-view training set. 

\section{Conclusion}

This paper focuses on the image-based localization (IBL) task which aims to estimate the geographical location of a query image by recognizing its nearest reference images from a city-scale dataset. 
We propose to tackle the problem of weak and noisy supervisions by self-supervising image-to-region similarities. 
Our method outperforms state-of-the-arts on the standard localization benchmarks by noticeable margins.

\subsection*{Acknowledgements}
This work is supported in part by SenseTime Group Limited, in part by the General Research Fund through the Research Grants Council of Hong Kong under Grants CUHK 14202217 / 14203118 / 14205615 / 14207814 / 14213616 / 14208417 / 14239816, in part by CUHK Direct Grant.

\clearpage
%
%
\bibliographystyle{splncs04}
\bibliography{2191}

\begin{thebibliography}{10}
\providecommand{\url}[1]{\texttt{#1}}
\providecommand{\urlprefix}{URL }
\providecommand{\doi}[1]{https://doi.org/#1}

\bibitem{anguelov2010google}
Anguelov, D., Dulong, C., Filip, D., Frueh, C., Lafon, S., Lyon, R., Ogale, A.,
  Vincent, L., Weaver, J.: Google street view: Capturing the world at street
  level. Computer  \textbf{43}(6),  32--38 (2010)

\bibitem{arandjelovic2016netvlad}
Arandjelovic, R., Gronat, P., Torii, A., Pajdla, T., Sivic, J.: Netvlad: Cnn
  architecture for weakly supervised place recognition. In: Proceedings of the
  IEEE conference on computer vision and pattern recognition. pp. 5297--5307
  (2016)

\bibitem{arandjelovic2013all}
Arandjelovic, R., Zisserman, A.: All about vlad. In: Proceedings of the IEEE
  conference on Computer Vision and Pattern Recognition. pp. 1578--1585 (2013)

\bibitem{arandjelovic2014dislocation}
Arandjelovi{\'c}, R., Zisserman, A.: Dislocation: Scalable descriptor
  distinctiveness for location recognition. In: Asian Conference on Computer
  Vision. pp. 188--204. Springer (2014)

\bibitem{ncarlevaris-2015a}
Carlevaris-Bianco, N., Ushani, A.K., Eustice, R.M.: University of {Michigan}
  {North} {Campus} long-term vision and lidar dataset. International Journal of
  Robotics Research  \textbf{35}(9),  1023--1035 (2015)

\bibitem{caron2018deep}
Caron, M., Bojanowski, P., Joulin, A., Douze, M.: Deep clustering for
  unsupervised learning of visual features. In: Proceedings of the European
  Conference on Computer Vision. pp. 132--149 (2018)

\bibitem{castle2008video}
Castle, R., Klein, G., Murray, D.W.: Video-rate localization in multiple maps
  for wearable augmented reality. In: 2008 12th IEEE International Symposium on
  Wearable Computers. pp. 15--22. IEEE (2008)

\bibitem{chen2011city}
Chen, D.M., Baatz, G., K{\"o}ser, K., Tsai, S.S., Vedantham, R.,
  Pylv{\"a}n{\"a}inen, T., Roimela, K., Chen, X., Bach, J., Pollefeys, M.,
  et~al.: City-scale landmark identification on mobile devices. In: CVPR 2011.
  pp. 737--744. IEEE (2011)

\bibitem{deng2009imagenet}
Deng, J., Dong, W., Socher, R., Li, L.J., Li, K., Fei-Fei, L.: Imagenet: A
  large-scale hierarchical image database. In: 2009 IEEE conference on computer
  vision and pattern recognition. pp. 248--255. Ieee (2009)

\bibitem{doersch2015unsupervised}
Doersch, C., Gupta, A., Efros, A.A.: Unsupervised visual representation
  learning by context prediction. In: Proceedings of the IEEE International
  Conference on Computer Vision. pp. 1422--1430 (2015)

\bibitem{fischler1981random}
Fischler, M.A., Bolles, R.C.: Random sample consensus: a paradigm for model
  fitting with applications to image analysis and automated cartography.
  Communications of the ACM  \textbf{24}(6),  381--395 (1981)

\bibitem{furlanello2018born}
Furlanello, T., Lipton, Z.C., Tschannen, M., Itti, L., Anandkumar, A.: Born
  again neural networks. International Conference on Machine Learning  (2018)

\bibitem{ge2020mutual}
Ge, Y., Chen, D., Li, H.: Mutual mean-teaching: Pseudo label refinery for
  unsupervised domain adaptation on person re-identification. In: International
  Conference on Learning Representations (2020)

\bibitem{ge2020selfpaced}
Ge, Y., Chen, D., Zhu, F., Zhao, R., Li, H.: Self-paced contrastive learning
  with hybrid memory for domain adaptive object re-id (2020)

\bibitem{ge2018fd}
Ge, Y., Li, Z., Zhao, H., Yin, G., Yi, S., Wang, X., Li, H.: Fd-gan:
  Pose-guided feature distilling gan for robust person re-identification. In:
  Advances in Neural Information Processing Systems. pp. 1229--1240 (2018)

\bibitem{ge2020structured}
Ge, Y., Zhu, F., Zhao, R., Li, H.: Structured domain adaptation with online
  relation regularization for unsupervised person re-id (2020)

\bibitem{gidaris2018unsupervised}
Gidaris, S., Singh, P., Komodakis, N.: Unsupervised representation learning by
  predicting image rotations. arXiv preprint arXiv:1803.07728  (2018)

\bibitem{goyal2019scaling}
Goyal, P., Mahajan, D., Gupta, A., Misra, I.: Scaling and benchmarking
  self-supervised visual representation learning. In: Proceedings of the IEEE
  International Conference on Computer Vision. pp. 6391--6400 (2019)

\bibitem{hane20173d}
H{\"a}ne, C., Heng, L., Lee, G.H., Fraundorfer, F., Furgale, P., Sattler, T.,
  Pollefeys, M.: 3d visual perception for self-driving cars using a
  multi-camera system: Calibration, mapping, localization, and obstacle
  detection. Image and Vision Computing  \textbf{68},  14--27 (2017)

\bibitem{hartley2003multiple}
Hartley, R., Zisserman, A.: Multiple view geometry in computer vision.
  Cambridge university press (2003)

\bibitem{hongsuck2018cplanet}
Hongsuck~Seo, P., Weyand, T., Sim, J., Han, B.: Cplanet: Enhancing image
  geolocalization by combinatorial partitioning of maps. In: Proceedings of the
  European Conference on Computer Vision. pp. 536--551 (2018)

\bibitem{jegou2008hamming}
Jegou, H., Douze, M., Schmid, C.: Hamming embedding and weak geometric
  consistency for large scale image search. In: European conference on computer
  vision. pp. 304--317. Springer (2008)

\bibitem{jegou2010aggregating}
J{\'e}gou, H., Douze, M., Schmid, C., P{\'e}rez, P.: Aggregating local
  descriptors into a compact image representation. In: 2010 IEEE computer
  society conference on computer vision and pattern recognition. pp.
  3304--3311. IEEE (2010)

\bibitem{jegou2011aggregating}
Jegou, H., Perronnin, F., Douze, M., S{\'a}nchez, J., Perez, P., Schmid, C.:
  Aggregating local image descriptors into compact codes. IEEE transactions on
  pattern analysis and machine intelligence  \textbf{34}(9),  1704--1716 (2011)

\bibitem{kim2017learned}
Kim, H.J., Dunn, E., Frahm, J.M.: Learned contextual feature reweighting for
  image geo-localization. In: 2017 IEEE Conference on Computer Vision and
  Pattern Recognition. pp. 3251--3260. IEEE (2017)

\bibitem{knopp2010avoiding}
Knopp, J., Sivic, J., Pajdla, T.: Avoiding confusing features in place
  recognition. In: European Conference on Computer Vision. pp. 748--761.
  Springer (2010)

\bibitem{kolesnikov2019revisiting}
Kolesnikov, A., Zhai, X., Beyer, L.: Revisiting self-supervised visual
  representation learning. In: Proceedings of the IEEE conference on Computer
  Vision and Pattern Recognition. pp. 1920--1929 (2019)

\bibitem{lecun2015deep}
LeCun, Y., Bengio, Y., Hinton, G.: Deep learning. nature  \textbf{521}(7553),
  436--444 (2015)

\bibitem{liu2017efficient}
Liu, L., Li, H., Dai, Y.: Efficient global 2d-3d matching for camera
  localization in a large-scale 3d map. In: Proceedings of the IEEE
  International Conference on Computer Vision. pp. 2372--2381 (2017)

\bibitem{liu2019stochastic}
Liu, L., Li, H., Dai, Y.: Stochastic attraction-repulsion embedding for large
  scale image localization. In: Proceedings of the IEEE International
  Conference on Computer Vision. pp. 2570--2579 (2019)

\bibitem{lowe2004distinctive}
Lowe, D.G.: Distinctive image features from scale-invariant keypoints.
  International journal of computer vision  \textbf{60}(2),  91--110 (2004)

\bibitem{RobotCarDatasetIJRR}
Maddern, W., Pascoe, G., Linegar, C., Newman, P.: {1 Year, 1000km: The Oxford
  RobotCar Dataset}. The International Journal of Robotics Research
  \textbf{36}(1),  3--15 (2017)

\bibitem{mur2015orb}
Mur-Artal, R., Montiel, J.M.M., Tardos, J.D.: Orb-slam: a versatile and
  accurate monocular slam system. IEEE transactions on robotics
  \textbf{31}(5),  1147--1163 (2015)

\bibitem{noroozi2016unsupervised}
Noroozi, M., Favaro, P.: Unsupervised learning of visual representations by
  solving jigsaw puzzles. In: European Conference on Computer Vision. pp.
  69--84. Springer (2016)

\bibitem{noroozi2017representation}
Noroozi, M., Pirsiavash, H., Favaro, P.: Representation learning by learning to
  count. In: Proceedings of the IEEE International Conference on Computer
  Vision. pp. 5898--5906 (2017)

\bibitem{paulin2015local}
Paulin, M., Douze, M., Harchaoui, Z., Mairal, J., Perronin, F., Schmid, C.:
  Local convolutional features with unsupervised training for image retrieval.
  In: Proceedings of the IEEE international conference on computer vision. pp.
  91--99 (2015)

\bibitem{perronnin2010large}
Perronnin, F., Liu, Y., S{\'a}nchez, J., Poirier, H.: Large-scale image
  retrieval with compressed fisher vectors. In: 2010 IEEE Computer Society
  Conference on Computer Vision and Pattern Recognition. pp. 3384--3391. IEEE
  (2010)

\bibitem{philbin2007object}
Philbin, J., Chum, O., Isard, M., Sivic, J., Zisserman, A.: Object retrieval
  with large vocabularies and fast spatial matching. In: 2007 IEEE conference
  on computer vision and pattern recognition. pp.~1--8. IEEE (2007)

\bibitem{Philbin08}
Philbin, J., Chum, O., Isard, M., Sivic, J., Zisserman, A.: Lost in
  quantization: {I}mproving particular object retrieval in large scale image
  databases. In: IEEE Conference on Computer Vision and Pattern Recognition
  (2008)

\bibitem{radenovic2016cnn}
Radenovi{\'c}, F., Tolias, G., Chum, O.: Cnn image retrieval learns from bow:
  Unsupervised fine-tuning with hard examples. In: European conference on
  computer vision. pp. 3--20. Springer (2016)

\bibitem{sattler2011fast}
Sattler, T., Leibe, B., Kobbelt, L.: Fast image-based localization using direct
  2d-to-3d matching. In: 2011 International Conference on Computer Vision. pp.
  667--674. IEEE (2011)

\bibitem{schindler2007city}
Schindler, G., Brown, M., Szeliski, R.: City-scale location recognition. In:
  2007 IEEE Conference on Computer Vision and Pattern Recognition. pp.~1--7.
  IEEE (2007)

\bibitem{shotton2013scene}
Shotton, J., Glocker, B., Zach, C., Izadi, S., Criminisi, A., Fitzgibbon, A.:
  Scene coordinate regression forests for camera relocalization in rgb-d
  images. In: Proceedings of the IEEE Conference on Computer Vision and Pattern
  Recognition. pp. 2930--2937 (2013)

\bibitem{simonyan2014very}
Simonyan, K., Zisserman, A.: Very deep convolutional networks for large-scale
  image recognition. arXiv preprint arXiv:1409.1556  (2014)

\bibitem{torii201524}
Torii, A., Arandjelovic, R., Sivic, J., Okutomi, M., Pajdla, T.: 24/7 place
  recognition by view synthesis. In: Proceedings of the IEEE Conference on
  Computer Vision and Pattern Recognition. pp. 1808--1817 (2015)

\bibitem{torii2013visual}
Torii, A., Sivic, J., Pajdla, T., Okutomi, M.: Visual place recognition with
  repetitive structures. In: Proceedings of the IEEE conference on computer
  vision and pattern recognition. pp. 883--890 (2013)

\bibitem{ullman1979interpretation}
Ullman, S.: The interpretation of structure from motion. Proceedings of the
  Royal Society of London. Series B. Biological Sciences  \textbf{203}(1153),
  405--426 (1979)

\bibitem{vo2017revisiting}
Vo, N., Jacobs, N., Hays, J.: Revisiting im2gps in the deep learning era. In:
  Proceedings of the IEEE International Conference on Computer Vision. pp.
  2621--2630 (2017)

\bibitem{weyand2016planet}
Weyand, T., Kostrikov, I., Philbin, J.: Planet-photo geolocation with
  convolutional neural networks. In: European Conference on Computer Vision.
  pp. 37--55. Springer (2016)

\bibitem{xie2020self}
Xie, Q., Hovy, E., Luong, M.T., Le, Q.V.: Self-training with noisy student
  improves imagenet classification. Proceedings of the IEEE Conference on
  Computer Vision and Pattern Recognition  (2020)

\bibitem{zeiler2014visualizing}
Zeiler, M.D., Fergus, R.: Visualizing and understanding convolutional networks.
  In: European conference on computer vision. pp. 818--833. Springer (2014)

\bibitem{zhang2020discriminability}
Zhang, M., Song, G., Zhou, H., Liu, Y.: Discriminability distillation in group
  representation learning. In: European Conference on Computer Vision (2020)

\bibitem{zhou2020rotate}
Zhou, H., Liu, J., Liu, Z., Liu, Y., Wang, X.: Rotate-and-render: Unsupervised
  photorealistic face rotation from single-view images. In: Proceedings of the
  IEEE/CVF Conference on Computer Vision and Pattern Recognition. pp.
  5911--5920 (2020)

\end{thebibliography}

\appendix

\section{Related Work (Cont.)}

\noindent\textbf{Image-based Localization (IBL).}
Besides the previous weakly-supervised methods \cite{arandjelovic2016netvlad,kim2017learned,liu2019stochastic} and our work,
there exists another stream of fully-supervised research which focused on datasets with 6DoF camera pose information \cite{ncarlevaris-2015a,chen2011city,RobotCarDatasetIJRR,shotton2013scene}.
Such datasets are generally difficult to collect and scale up, since they require extra human and computational costs to capture dense images and obtain 6DoF information via post-processing, \eg mapping, SfM \cite{ullman1979interpretation}, \textit{etc}.
In contrast, datasets \cite{torii2013visual,torii201524} studied in weakly-supervised methods are much easier to scale up, since the data and GPS information can be easily obtained for free from the internet without extra costs, \eg Google Street View \cite{anguelov2010google}, Baidu Total View.

\section{Progressively Refined Supervisions}

As shown in Tab. \ref{tab:gen}, the retrieval accuracies gradually increase as the network generation proceeds and saturate after the $4^\text{th}$ one, which indicates that the self-predicted soft supervisions in our work are progressively refined by training in generations.

\begin{table}[H]
	\centering
	\caption{Performance of our proposed method in different generations on Tokyo 24/7, in terms of Recall@1/5/10 (\%)}
	\label{tab:gen}
	\begin{center}
	\begin{tabular}{C{2cm}|C{1.5cm}C{1.5cm}C{1.5cm}C{1.5cm}}
	\hline
	Metric & $1^\text{st}$ & $2^\text{nd}$ & $3^\text{rd}$ & \textbf{$4^\text{th}$}  \\ 
	\hline 
	R$@1$ & 80.6 & 82.6 & 84.2 & \textbf{85.4} \\
	R$@5$& 87.6 & 89.2 & 90.5 & \textbf{91.1} \\
	R$@10$ &90.8 &92.7 & 92.8 & \textbf{93.3 }\\
	\hline
	\end{tabular}
	\end{center}
\end{table}

\end{document}